\documentclass[lettersize,journal]{IEEEtran}

\usepackage{graphicx}
\usepackage[colorlinks,linkcolor=black,anchorcolor=black,citecolor=black]{hyperref}
\usepackage{lineno,hyperref}
\modulolinenumbers[5]
\usepackage[cmex10]{amsmath}
\usepackage{amsfonts}

\usepackage{color}
\usepackage{algorithm}
\usepackage{algorithmic}
\usepackage{array}
\usepackage{multirow}
\usepackage{booktabs}
\usepackage{subfigure}
\usepackage{bm}
\usepackage{graphicx}
\usepackage{cite}
\usepackage{makecell}
\usepackage{amsthm}
\usepackage{amsmath}
\DeclareMathOperator{\Sim}{Sim}
\newtheorem{remark}{Remark}

\allowdisplaybreaks[4]

\usepackage{multirow}

\usepackage{amsmath}
\usepackage{stfloats}
\usepackage{float}
\usepackage{lipsum}

\makeatletter
\def\subsubsection{\@startsection{subsubsection}{3}{\parindent}{0.5ex plus 1ex minus 0.1ex}{0pt}{\normalfont\normalsize}}

\begin{document}
\title{Knowledge Graph Based Explainable and Generalized Zero-Shot Semantic Communications}

\author{Zhaoyu Zhang, Lingyi Wang,
Wei Wu,~\IEEEmembership{Member,~IEEE,}\\
Fuhui Zhou,~\IEEEmembership{Senior Member,~IEEE,}
Qihui Wu,~\IEEEmembership{Fellow,~IEEE}

\thanks{This work was supported by 
the National Natural Science Foundation of China under Grant 62271267, 
the open research fund of National Mobile Communications Research Laboratory, Southeast University under Grant 2024D16
and the key program of Marine Economy Development Special Foundation of Department of Natural Resources of Guangdong Province (GDNRC[2023]24).
(Corresponding authors: Wei Wu and Fuhui Zhou)}
\thanks{Zhaoyu Zhang is with the School of Computer Science,
Nanjing University of Posts and Telecommunications, Nanjing, 210003, China.
(e-mail: b22040502@njupt.edu.cn).}
\thanks{Lingyi Wang is with the College of Science,
Nanjing University of Posts and Telecommunications, Nanjing, 210003, China.
(e-mail: lingyiwang@njupt.edu.cn).}
\thanks{
Wei Wu is with the College of Communication and Information Engineering,
Nanjing University of Posts and Telecommunications, Nanjing, 210003, China,
and also with the National Mobile Communications Research Laboratory, Southeast University, Nanjing, 210096, China.
(e-mail: weiwu@njupt.edu.cn).}
\thanks{Fuhui Zhou and Qihui Wu are with the College of Electronic and Information Engineering, 
Nanjing University of Aeronautics and Astronautics, Nanjing, 210000, China.
(e-mail: zhoufuhui@ieee.org, wuqihui2014@sina.com).}
}

\maketitle

\begin{abstract}
  Data-driven semantic communication is based on superficial statistical patterns, thereby lacking interpretability and generalization, especially for applications with the presence of unseen data.
  To address these challenges, we propose a novel knowledge graph-enhanced zero-shot semantic communication (KGZS-SC) network. 
  Guided by the structured semantic information from a knowledge graph-based semantic knowledge base (KG-SKB), our scheme provides generalized semantic representations and enables reasoning for unseen cases.
  Specifically, the KG-SKB aligns the semantic features in a shared category semantics embedding space and enhances the generalization ability of the transmitter through aligned semantic features, 
  thus reducing communication overhead by selectively transmitting compact visual semantics. 
  At the receiver, zero-shot learning (ZSL) is leveraged to enable direct classification for unseen cases without the demand for retraining or additional computational overhead, 
  thereby enhancing the adaptability and efficiency of the classification process in dynamic or resource-constrained environments. 
  The simulation results conducted on the APY datasets show that the proposed KGZS-SC network exhibits robust generalization and significantly outperforms existing SC frameworks in classifying unseen categories across a range of SNR levels.
\end{abstract}  
  

\IEEEpeerreviewmaketitle

\section{Introduction}
Semantic communication has emerged as a critical paradigm for future intelligent communication systems, such as digital twins and virtual reality \cite{chaccour2024less}.
Different from the traditional syntactic communication paradigm that puts emphasis on symbol-level recovery accuracy, 
semantic communication paradigm aims to cognize, understand, and selectively convey the essential intention, thereby significantly reducing data redundancy and enhancing communication efficiency \cite{10538233,10158526}. 
To extract the critical features from original data, deep learning (DL) has been widely explored due to its powerful representation ability \cite{10622764}.
However, the majority of existing DL-enabled semantic communication networks are formulated in a data-driven manner \cite{xie2021deep}, relying solely on superficial statistical patterns rather than leveraging the inherent knowledge embedded within the data.
For the generalized intelligent communication applications required to handle unseen data or scenarios, data-driven semantic communication becomes unreliable and inapplicable due to its lack of interpretability and generalizability.

To address the aforementioned issues, the data-driven semantic communication can evolve into the knowledge-driven semantic communication by integrating knowledge graphs  \cite{9838470}. 
The knowledge graph is capable of capturing relationships and semantic information between entities through explicit structured representations,  
instead of typically relying on large datasets to learn patterns \cite{jiang2022reliable}.
Moreover, the knowledge graph organizes semantic information in the form of triples, defined by $\langle entity, relation, entity \rangle$, which inherently enhances the transparency, interpretability, and robustness of the communication process. 
The authors in \cite{9838470} explored inference rules through the knowledge graph for robust semantic communication, where a semantic correction algorithm was designed to rectify errors at the semantic level.
In \cite{10333452}, the probability graph was used to compress the unnecessary triples in the knowledge graph for more efficient resource utilization.
However, the information contained in a knowledge graph is limited, while dynamic, diverse scenarios for intelligent applications can involve a large number of unseen entities, relationships, or semantic patterns. 
Thus, the generalization ability of knowledge-driven semantic communication is important \cite{10183794}.

Zero-shot learning (ZSL) enables inference for unseen data by generalizing knowledge from previously encountered domains,
which is characterized by strong adaptability, cross-domain transferability, and significantly reduced training costs \cite{nayak2020zero,wang2018zero,kampffmeyer2019rethinking,sun2023semantic}.
While the work \cite{nayak2020zero,wang2018zero,kampffmeyer2019rethinking} explores efficient locally deployed zero-shot algorithms, 
zero-shot semantic communication needs to further tackle bandwidth and computational constraints in complex communication environments.
In \cite{sun2023semantic}, the authors proposed the wireless zero-shot object recognition with attribute-based representations.
However, although these representations can enhance recognition for structurally similar objects, they are hard to address the disjoint distributions.
Moreover, \cite{sun2023semantic} lacks the generalizability to explicitly infer the unseen entities and relationships, which reduces the transparency and interpretability of the semantic communications.

To the best of our knowledge, this paper is the first to propose an end-to-end knowledge graph-enhanced zero-shot semantic communication (KGZS-SC) framework, 
which aims to tackle the challenges of generalization, explainability, and robustness in semantic communication with unseen data. 
Particularly, we design a knowledge graph-based semantic knowledge base (KG-SKB) that embeds category-level semantics into a unified high-dimensional category semantics embedding space and exploits explicit graph-structured relations to align visual features with semantic embeddings, 
effectively mitigating the hubness problem in zero-shot learning. By integrating structured knowledge-graph information at the receiver, KGZS-SC supports real-time classification of novel classes with zero additional computational cost, thereby enhancing interpretability, generalization, and training efficiency. 
Extensive experiments on the APY dataset under varying SNR conditions demonstrate that the proposed KGZS-SC significantly outperforms the existing semantic communication schemes in classification accuracy and generalization to unseen categories.

\vspace{-0.4cm}
\section{System Model}
The semantic system consists of a single-antenna transmitter, a physical channel, and a single-antenna receiver. 
The transmitter integrates a semantic encoder and a channel encoder, while both the transmitter and receiver share access to a unified semantic knowledge base (SKB). 
At the receiver, a channel decoder and a semantic decoder collaborate to infer the communication intention from semantic symbols and execute the semantic tasks.
We consider a practical scenario that the data at the transmitter is mixed seen and unseen. 

The transmitter processes the raw observation data by extracting the critical semantic features.
Given the observation $\boldsymbol{x}$, the semantic symbol $\boldsymbol{z}$ can be obtained by
$
  \boldsymbol{z} = H_{\beta}(S_{\alpha}(\boldsymbol{x})),
$
where $S_{\alpha}(\cdot)$ and $H_{\beta}(\cdot)$ are respectively 
the $\alpha$-parameterized semantic encoder and the $\beta$-parameterized channel encoder.
The semantic symbol $\boldsymbol{z}$ is transmitted over the physical wireless channel in practice. To enable the end-to-end framework of semantic coding, the received semantic symbol is obtained through a simulated channel during the training, represented by
$
\hat{\boldsymbol{z}} = h \boldsymbol{z} + \boldsymbol{n},
$
where $h \in [0,1]$ represents the channel gain, and
\( \boldsymbol{n} \sim \mathcal{N}(0, \sigma^2\mathbf{I}) \) represents the additive white Gaussian noise (AWGN).
The receiver decodes the semantic symbol $\hat{\boldsymbol{z}}$ and complete the semantic communication.
The critical semantics $\hat{\boldsymbol{s}}$ for the task completion is represented by
$
\hat{\boldsymbol{s}} = S^{-1}_{\hat{\beta}}(H^{-1}_{\hat{\alpha}}(\hat{\boldsymbol{z}})),
$
where $H^{-1}_{\hat{\beta}}(\cdot)$ and $S^{-1}_{\hat{\alpha}}(\cdot)$ are respectively 
the $\hat{\beta}$-parameterized channel decoder and the $\hat{\alpha}$-parameterized semantic decoder.

In real-world applications, unseen data necessitates the generalizability for semantic representation to capture the features of both seen and unseen features. 
In this study, we select the image classification task as a representative case study. 
Particularly, we divide the image categories into seen sets $\mathcal{Y}_s$ and unseen category $\mathcal{Y}_u$ based on the category, which satisfy $\mathcal{Y}_s \cap \mathcal{Y}_u = \emptyset$ and $\mathcal{Y}_s \cup \mathcal{Y}_u = \mathcal{Y}$,  $\mathcal{Y} = \mathcal{Y}_s \cup \mathcal{Y}_u$. 
Both $\mathcal{Y}_s$ and $\mathcal{Y}_u$ share a unified semantic space defined by the SKB. The receiver is required to classify over $\mathcal{Y}$ based on the representations from the transmitter.
The seen data, serving as the training dataset, is represented as 
$\mathcal{S} = \{(\boldsymbol{x}_i^s, y_i^s)\}_{i=1}^{N_s}$ with $N_s$ training samples, where $x_i^s \in \mathcal{X}_s$ represents the $i$-th original image, 
and $y_i^s \in \mathcal{Y}_s$ corresponds to its associated category label. The unseen data from dynamic real-world scenarios is represented as 
$\mathcal{U} = \{(\boldsymbol{x}_i^u, y_i^u)\}_{i=1}^{N_u}$, where $N_u$ represents the number of unseen samples.

\begin{remark}
In real-world applications characterized by dynamic and diverse environments, the unseen data is inherently neither definitive nor finite. 
Here, the division of data into seen and unseen datasets is introduced primarily to facilitate a clear and structured description of our proposed framework.
  Both the unseen categories $\mathcal{Y}_u$ and the number of the unseen categories $|\mathcal{Y}_u|$ are unknown for the transceivers.
\end{remark}

\vspace{-0.5cm}
\section{Proposed KGZS-SC Framework}
In this section,  a novel KGZS-SC framework with KG-SKB, as shown in Fig. 1, is proposed to enhance the generalization and interpretability of semantic communications.

\begin{figure*}[htbp]
  \centering
  \includegraphics[width=0.95\textwidth]{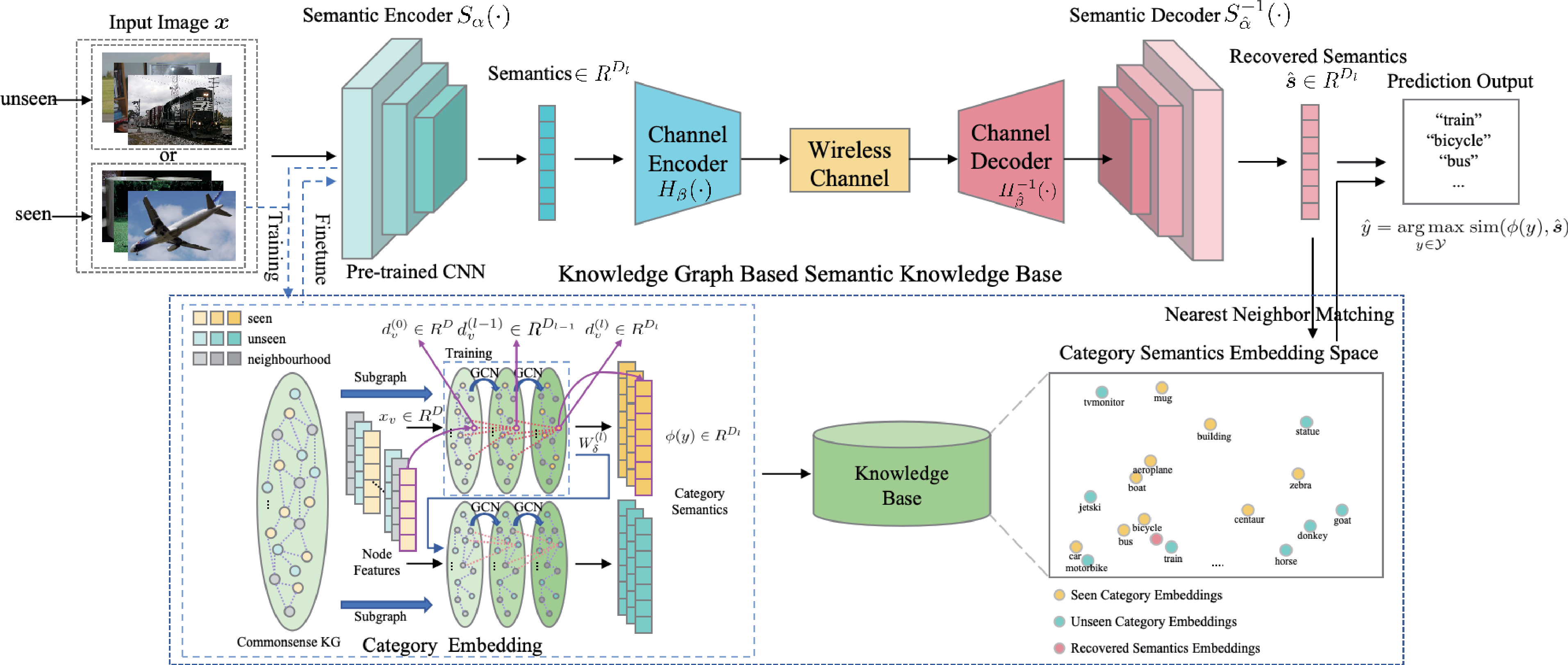}
  \vspace{-0.2cm}
  \caption{The proposed knowledge graph enhanced zero-shot semantic communication network with generalization ability.}
  \vspace{-0.4cm}
\end{figure*}

\vspace{-0.3cm}
\subsection{KG-SKB Initialization}
The graph for $y \in \mathcal{Y}_s$ is initialized from a two-hop subgraph in commonsense KG \cite{speer2017conceptnet}. Particularly, we first collect the one-hop neighborhood $N^1(y)$ and two-hop neighbors $N^2(y)$ of each node $y$ to form category subgraphs $\mathcal G_s=(\mathcal V_s,\mathcal E_s)$.  
The global graph $\mathcal G_S=(\mathcal V_S,\mathcal E_S)$ for seen categories is obtained by unifying the node sets and edge sets, represented by
\begin{equation}
  \mathcal V_S = \bigcup_{y\in \mathcal{Y}_s}\mathcal V_s,
  \quad
  \mathcal E_S = \bigcup_{y\in \mathcal{Y}_s}\mathcal E_s.
\end{equation}
Each node $v\in\mathcal V_S$ is initialized by searching pre-trained GloVe embeddings \cite{pennington2014glove}, represented by
\begin{equation}
  \boldsymbol{d}_v^{(0)} = \mathrm{GloVe}(v)\in R^D,
\end{equation}
where $D$ represents the dimensionality of node embeddings. Let $C_{uv}$ be the number of times a random walk starting at $u$ and arriving at node $v$. Then, the edge weights $s_{uv}$ from the node $u$ to the node $v$ is obtained by Laplace smoothing and rownormalization as
\begin{equation}
  \widetilde C_{uv} = C_{uv} + 1,
  \qquad
  s_{uv}
  = \frac{\widetilde C_{uv}}{\sum_{w\in\mathcal V_S}\widetilde C_{uw}}.
\end{equation}

\vspace{-0.5cm}
\subsection{GCN-Based Generalized Semantic Representation}
As shown in Fig. 1, a novel two-layer representation approach based on graph convolutional network (GCN) \cite{kipf2016semi} is further proposed to capture both local and global semantic dependencies from the initialized KG-SKB.  
To train GCN and obtain generalized semantic representations, each node \(v\) aggregates the features of its neighbors \(u\in N^1(v)\cup\{v\}\) by
\begin{equation}
  \boldsymbol{z}_v^{(l)}
  = \sum_{u}
    \frac{1}{m_{v u}}W_{\delta}^{(l)}\,\boldsymbol{d}_u^{(l-1)}, \, \,
  m_{v u} = \sqrt{\deg(v)\deg(u)},
\end{equation}
where \(\boldsymbol{d}_u^{(l-1)}\) is the feature of neighbor \(u\) at layer \(l-1\), \(W_{\delta}^{(l)}\) is the learnable weight, and \(m_{v u}\) is the symmetric normalization factor, and $\deg(v)$ represents the degrees of node $v$.
The residual structure with layer normalization is then used to preserve original information and mitigate over-smoothing by
\begin{equation}
  \tilde{\boldsymbol z}_v^{(l)}
  = \mathrm{LN}\bigl(\boldsymbol z_v^{(l)} + \boldsymbol d_v^{(l-1)}\bigr),
\end{equation}
\begin{equation}
  \mathrm{LN}(\boldsymbol k)
  = \frac{\boldsymbol k - \mu_{\boldsymbol k}}{\sqrt{\sigma_{\boldsymbol k}^2 + \epsilon}}\;\gamma + \eta,
\end{equation}
where \(\mu_{\boldsymbol k}\) and \(\sigma_{\boldsymbol k}^2\) are respectively the mean and variance of \(\boldsymbol k\), \(\epsilon\) is a small constant, $\gamma$ is the learnable scale, and $\eta$ is the shift parameter. Then, a non-linear activation is utilized to obtain the feature of node \(v\) at layer \(l\) as
\begin{equation}
  \boldsymbol d_v^{(l)}
  = \varphi\bigl(\tilde{\boldsymbol z}_v^{(l)}\bigr)\in R^{D_l},
\end{equation}
where \(\varphi(\cdot)\) is the ReLU function.  
The final representation $d_v^{(L)}\in R^{D_l}$ of node $v$ at the last layer $L$ is adopted as the semantic embedding $\phi(y)$ for the seen category $y \in \mathcal{Y}_s$.
For unseen categories, we extract and pre-process the unseen global subgraph \(\mathcal G_U\) in the same manner as the construction of \(\mathcal G_S\),  and then apply the trained GCN to obtain the unseen semantic category representations. Now, we have the semantic representations for all categories $\mathcal{Y} = \mathcal{Y}_s \cup \mathcal{Y}_u$, which are then used for subsequent similarity scoring and zero-shot inference.

\vspace{-0.3cm}
\subsection{Training Process And Loss Functions}
We propose a two-stage training process.
In stage one, the semantic encoder leverages the pre-trained ResNet to extract semantic features, ensuring a robust initialization for feature extraction. 
Meanwhile, the goal of the GCN is to maximize the similarity between the extracted semantic features and the correct category $y \in \mathcal{Y}_s$.
The loss function of stage one can be expressed as
\begin{align}
  \mathcal{L}_{1}(\alpha, \delta) = -\log \frac{\exp( \Sim(S_{\alpha}(\boldsymbol{x}), \phi(y)))}{\sum_{y' \in \mathcal{Y}_s} \exp( \Sim(S_{\alpha}(\boldsymbol{x}), \phi(y^\prime)))},
\end{align}
where $\alpha$ is the parameter of the semantic encoder \(S_\alpha\), $\delta$ is the parameter of the GCN, $y$ is the category label of the input $\boldsymbol{x} \in \mathcal{X}_s$, and \(\Sim(\mathbf{a},\mathbf{b}) = \| \mathbf{a} - \mathbf{b} \|_2 \) measures the  semantic similarity based on Euclidean distance.

In stage two, the channel encoder and decoder are fine-tuned. The loss function of stage two can be expressed as
\begin{align}
  \mathcal{L}_{2}(\beta, \hat{\beta}) = \underbrace{\|\hat{\boldsymbol{s}}-S_{\alpha}(\boldsymbol{x})\|_2^2}_{\text{Recovery Loss}} + \lambda \underbrace{\|\hat{\boldsymbol{s}} - \phi(y)\|_2^2}_{{\text{Alignment Loss}}},
\end{align}
where $\lambda$ represents the weight factor.  
The recovery loss term addresses the noise introduced by the physical channel, aiming to minimize the semantic distance between the original semantics and the recovered semantics. 
This process functions as a semantic-level channel feedback mechanism, ensuring robust communication.
On the other hand, the alignment loss term focuses on aligning the decoded semantics with the generalized category representations, which serves as reverse guidance provided by the GCN, 
reinforcing the semantic consistency across the communication pipeline.

In practice, the vector matching is used to determine the classification result $\hat{y}$, which can be expressed as
\begin{align}
  \hat{y} = \underset{y \in \mathcal{Y}}{\arg\max} \ \text{Sim}(\phi(y), \hat{\boldsymbol{s}}) \,= \underset{y \in \mathcal{Y}}{\arg\max} \, \| \phi(y) - \hat{\boldsymbol{s}}\|_2.
\end{align}

\vspace{-0.5cm}
\subsection{Complexity Analysis}
Let $V$ and $E$ respectively be the numbers of nodes and edges in the semantic knowledge graph, and $N_s$ be the number of training images.
The computational complexity of the KGZS-SC framework has three linearly scaling components: (1) the construction of the knowledge-graph semantic knowledge base, which requires a two-hop subgraph extraction and subsequent pruning and embedding initialization in $O(|\mathcal{Y}_s|(V+E) + V)$ time; 
(2) the joint training of the GCN and ResNet-based semantic encoder, whose per-epoch cost is $O\bigl(N_s\,(M + (V+E)D)\bigr)$ for stage one and $O\bigl(N_s\,(M+M')\bigr)$ for stage two, where $M$ and $M'$ denote the per-sample costs of the ResNet and the channel encoder/decoder respectively; 
and (3) zero-shot inference, which incurs $O(M + M' + |\mathcal{Y}|D)$ per image for feature extraction and class matching with $|\mathcal{Y}| = |\mathcal{Y}_s \cup \mathcal{Y}_u|$. Since the KG-SKB is built and optimized offline, the real-time overhead at the transmitter and receiver is from zero-shot inference in practice, which is $O(M + M' + |\mathcal{Y}|D)$.

\vspace{-0.5cm}
\section{Simulation Results}
In this section, we compare the proposed KGZS-SC with several baselines under the AWGN channels. 
We adopt APY as the dataset, which consists of 20 seen categories for training and 12 unseen categories for testing. The learning rate of both the semantic coding networks and GCNs is set to 0.0001, $\lambda = 0.9$ and $\epsilon = 0.001$.
The dimensional size of the node feature is 300 and both the category semantics and semantics is set to 2049.
We adopt the 16-QAM as the modulation method for transmission \cite{10622764} that is widely used in LTE and IEEE 802.11 deployments. 
We also evaluate the model performance in a practical setting with 70\% seen categories and 30\% unseen categories. Moreover, the harmonic mean metric $\xi$ \cite{nayak2020zero} is used to measure the generalization ability, which is calculated as
$
  \xi =\frac{2 \times \text { Seen Accuracy } \times \text { Unseen Accuracy }}{\text { Seen Accuracy }+ \text { Unseen Accuracy }}.
$
For better comparison, we introduce three baselines: (1) The basic deep learning-enabled semantic communication (DL-SC) \cite{weng2023deep}, (2) The zero-shot semantic communication (ZSL-SC) \cite{sun2023semantic} without KG, and (3) The knowledge graph based semantic communication (KG-SC) \cite{9838470} without zero-shot learning. 

\begin{figure}[t]
  \centering
  \includegraphics[scale=0.3]{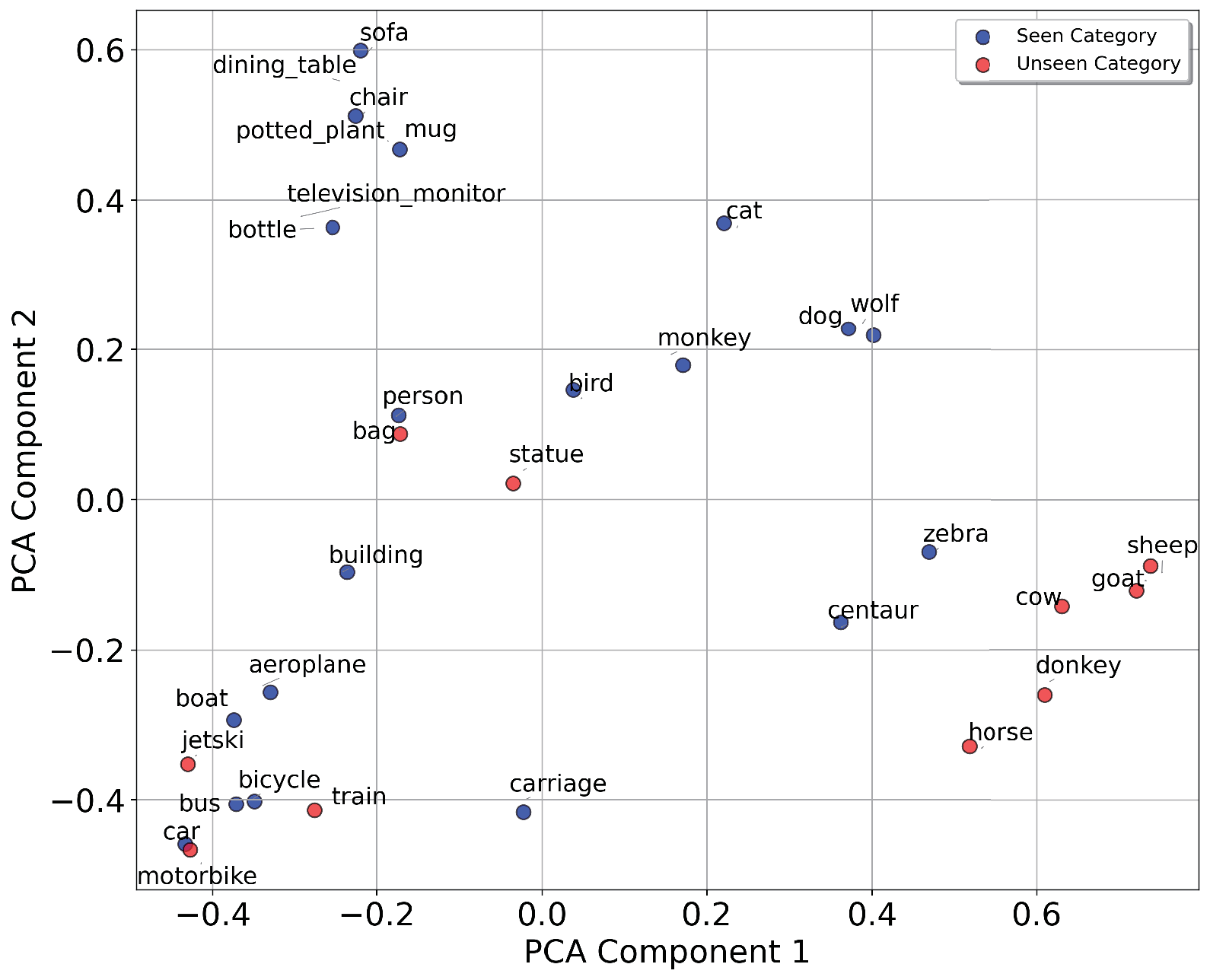} 
  \vspace{-0.3cm}
  \caption{The PCA visualization of the proposed KGZS-SC.}
  \vspace{-0.6cm}
\end{figure}
Fig. 2 shows the principal component analysis (PCA) of the proposed KGZS-SC, which reveals the extent of generalization from seen to unseen categories. The x-axis and y-axis represent the first and second principal components of the feature space, capturing the most significant variations in category embeddings, where proximity indicates feature similarity and potential knowledge transfer effectiveness in ZSL.
Unseen categories positioned near semantically similar seen categories, such as ``zebra'' near ``cow'' or ``jetski'' near ``boat'', suggest that the model successfully captures shared attributes, enabling effective knowledge transfer. However, isolated unseen categories like ``statue'' or ``donkey'' indicate that the learned feature space lacks sufficient semantic grounding for certain novel categories, limiting generalization. 
This disparity highlights a key challenge in ZSL: while attribute-based representations can facilitate recognition for structurally similar objects, they are hard to address the disjoint distributions. By enhancing the knowledge transfer based on knowledge graph based semantic embeddings with structured information from external sources, our proposed KGZS-SC can significantly improve the ZSL performance for semantic communication by reducing the gap between seen and unseen representations.

\begin{figure*}[h!]
  \centering
  \includegraphics[scale=0.5]{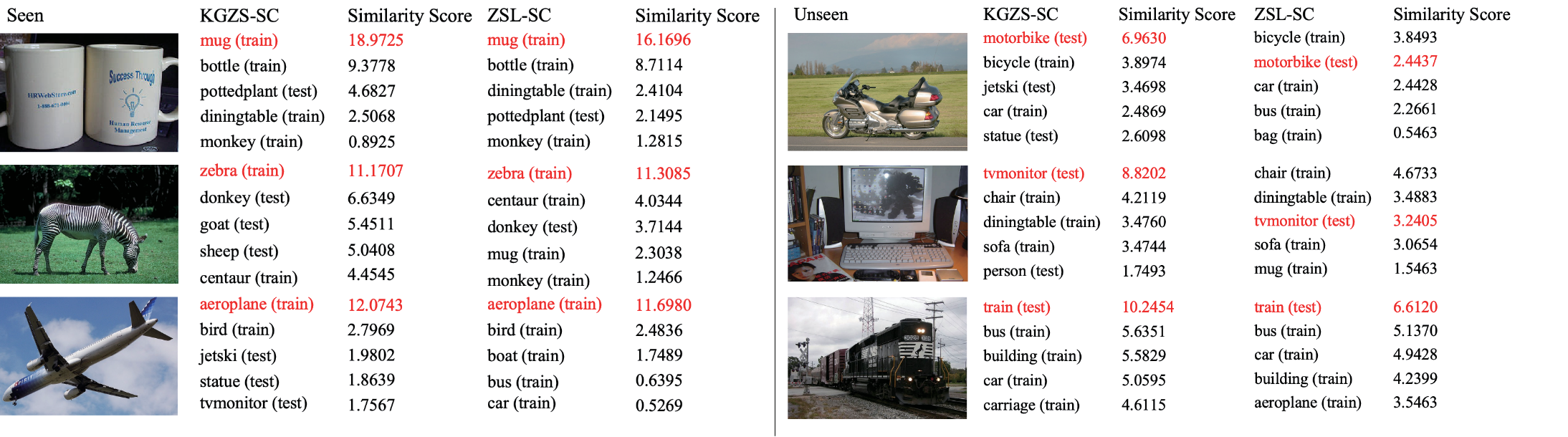} 
  \vspace{-0.3cm}
  \caption{The classification visualization results and the similarity score comparison of the ZSL schemes.}
  \vspace{-0.3cm}
\end{figure*}
Fig. 3 illustrates the classification visualization results and the similarity score comparison of the ZSL schemes. The visualization reveals a striking disparity between KGZS-SC and ZSL-SC in ZSL performance, particularly in their ability to transfer knowledge to unseen categories. KGZS-SC maintains relatively high confidence scores for both seen and unseen categories, demonstrating a more balanced feature space that enables effective generalization. 
In contrast, ZSL-SC exhibits a severe confidence drop for unseen categories, indicating that its learned representations are heavily biased toward seen classes and fail to transfer knowledge effectively. The significantly higher scores for unseen categories in KGZS-SC suggest that incorporating structured knowledge from the knowledge graph, enhances semantic consistency across domains. 
This contrast underscores the limitation of standard ZSL approaches, where the absence of explicit relational knowledge leads to weak semantic associations for unseen classes. KGZS-SC mitigates this issue by integrating external knowledge, reducing the generalization gap and improving model robustness.

\begin{table*}[!htbp]
  \centering
  \caption{Performance comparison of the proposed KGZS-SC with various baselines on APY dataset.}
  \vspace{-0.2cm}
  \begin{tabular}{|c|c|c|c|c|c|}
  \hline
  Approach & SNR & Accuracy (Seen Category) & Accuracy (Unseen Category) & Harmonic Mean & Accuracy (Practical Application) \\ \hline
  \multirow{6}{*}{\makecell{KGZS-SC \\ (\textbf{Ours})}}
  & -10 & $57.43 \pm 1.32 $ & $41.64 \pm 1.21 $ & $48.28 \pm 1.14 $ & $52.69 \pm 1.12$ \\
  & -5 & $66.12 \pm 1.24 $ & $50.32 \pm 1.14 $ & $57.15 \pm 1.12 $ & $61.38 \pm 0.83$ \\ 
  & 0  & $67.58 \pm 1.12 $ & $52.70 \pm 1.10$ & $59.22 \pm 0.87 $ & $63.12 \pm 0.54$\\ 
  & 5  & $68.96 \pm 0.50 $ & $54.01 \pm 0.53 $ & $60.58 \pm 0.53 $ & $64.48 \pm 0.21$\\ 
  & 10 & $69.65 \pm 0.50$ & $53.33 \pm 0.55$ & $60.41 \pm 0.32 $ & $64.75 \pm 0.23$\\
  & 15 & $69.72 \pm 0.53$ & $53.81 \pm 0.34 $ & $60.74 \pm 0.30 $ & $64.95 \pm 0.22$\\ \hline
  \multirow{6}{*}{DL-SC} 
  & -10 & $50.63 \pm 1.12 $ & $-$ & $-$ & $35.44 \pm 0.78$\\
  & -5 & $57.82 \pm 1.19 $ & $-$ & $-$ & $40.47 \pm 0.83$\\ 
  & 0  & $61.60 \pm 0.88 $ & $-$ & $-$ & $43.12 \pm 0.62$\\ 
  & 5  & $68.74 \pm 0.62 $ & $-$ & $-$ & $48.12 \pm 0.43$\\ 
  & 10 & $73.41 \pm 0.30 $ & $-$ & $-$ & $51.39 \pm 0.21$\\
  & 15 & $76.87 \pm 0.21$ & $-$ & $-$ & $53.81 \pm 0.15$\\ \hline
  \multirow{6}{*}{ZSL-SC} 
  & -10 & $33.28 \pm 1.04 $ & $27.92 \pm 1.19 $ & $30.37 \pm 1.15 $ & $31.67 \pm 1.17$ \\
  & -5 & $36.53 \pm 1.25 $ & $27.06 \pm 1.17 $ & $31.09 \pm 0.92 $ & $33.69 \pm 1.23$\\ 
  & 0  & $51.76 \pm 0.89 $ & $28.77 \pm 0.74$ & $36.98 \pm 0.78$ & $44.86 \pm 0.64$\\ 
  & 5  & $61.35 \pm 0.79 $ & $29.68 \pm 0.66$ & $40.01 \pm 0.76$ & $51.85 \pm 0.65$\\ 
  & 10 & $62.83 \pm 0.57$ & $30.93 \pm 0.49$ & $41.45 \pm 0.54$ & $53.26 \pm 0.52$\\
  & 15 & $65.34 \pm 0.38$ & $34.68 \pm 0.26$ & $45.31 \pm 0.25$ & $56.14 \pm 0.29$\\ \hline
  \multirow{6}{*}{KG-SC} 
  & -10 & $58.10 \pm 0.80$ & $-$ & $-$ & $40.67 \pm 0.56$\\
  & -5 & $60.22 \pm0.81 $ & $-$ & $-$ & $42.15 \pm 0.57$\\ 
  & 0  & $62.43 \pm 0.73$ & $-$ & $-$ & $43.70 \pm 0.51$\\ 
  & 5  & $70.31 \pm 0.49$ & $-$ & $-$ & $49.22 \pm 0.35$\\ 
  & 10 & $74.26 \pm 0.48$ & $-$ & $-$ & $51.98 \pm 0.35$\\
  & 15 & $78.67 \pm 0.34$ & $-$ & $-$ & $55.07 \pm 0.23$\\ \hline
  \end{tabular} 
  \vspace{-0.4cm}
\end{table*}
  Table I demonstrates the performance comparison of the proposed KGZS-SC with baselines on the APY dataset in the classification accuracy, where different SNR reflects different noise strengths over the wireless channel. 
  As SNR decreases, the AWGN channel exhibits higher bit-error rates at the physical layer. 
  The mean and variance of the accuracy are obtained by averaging over 100 test episodes. 
  It is observed that the proposed KGZS-SC significantly outperforms the other benchmark schemes, with the 72.9\%, 49.1\%, 59.1\% and 15.7\% improvement in the classification accuracy in seen categories and unseen categories, harmonic mean and the classification accuracy in the practical application compared to the ZSL-SC scheme at the SNR of -10. 
  This is because the proposed knowledge graph-enhanced method can enable the semantic communication network with robustness against the physical noise by capturing both local and global semantic dependencies, thus providing the reliable knowledge representation. Hence, the proposed scheme significantly improves the semantic task performance even at the low SNR. 
  At high SNR, the channel noise is negligible and all models are close to near-ceiling semantic accuracy of neural networks. 
  Moreover, the traditional semantic communication schemes without zero-shot learning methods, such as DL-SC and KG-SC, fail to cognize the unseen categories. It demonstrates that DL-SC and KG-SC do not have the generalization ability, which is important for real-world communication applications.

\vspace{-0.5cm}
\section{Conclusion}
\vspace{-0.4cm}
To address the challenges of explainable and robust semantic communication for generalized intelligent applications with unseen data or scenarios, 
we have proposed the novel KGZS-SC framework, where the KG-SKB can embed category-level semantics into a shared high-dimensional category semantics embedding space and establish explicit relationships between visual and semantic features. 
By leveraging structured semantic information from the knowledge graph, the framework demonstrated improved interpretability, 
generalization, and efficiency, enabling direct classification of unseen categories without retraining or additional computational overhead. 
Simulation results have demonstrated that the proposed KGZS-SC exhibits superior generalization performance with unseen data compared to existing semantic communication frameworks.

\bibliography{IEEEabrv,mybibfile.bib}

\end{document}